\documentclass[letterpaper, 10 pt, conference]{ieeeconf}  % Comment this line out if you need a4paper

\IEEEoverridecommandlockouts                              % This command is only needed if 
\usepackage{amsmath} % assumes amsmath package installed
\usepackage{amssymb}  % assumes amsmath package installed

\title{\LARGE \bf
Efficient Optimal Path Planning in Dynamic Environments Using Koopman MPC}

\author{Mohammad Abtahi$^{1}$, Navid Mojahed$^{1}$, and Shima Nazari$^{1}$% <-this % stops a space
\thanks{$^{1}$The authors are with the Department of Mechanical and Aerospace Engineering, University of California Davis, USA (e-mail: sabtahi@ucdavis.edu; nmojahed@ucdavis.edu;  snazari@ucdavis.edu)}%
}

\usepackage{graphicx}
\usepackage{amsmath}
\usepackage{booktabs}
\usepackage{xcolor}

\begin{document}

\maketitle
\thispagestyle{empty}
\pagestyle{empty}

%%%%%%%%%%%%%%%%%%%%%%%%%%%%%%%%%%%%%%%%%%%%%%%%%%%%%%%%%%%%%%%%%%%%%%%%%%%%%%%%
\begin{abstract}
This paper presents a data-driven model predictive control framework for mobile robots navigating in dynamic environments, leveraging Koopman operator theory. Unlike the conventional Koopman-based approaches that focus on the linearization of system dynamics only, our work focuses on finding a global linear representation for the optimal path planning problem that includes both the nonlinear robot dynamics and collision-avoidance constraints. We deploy extended dynamic mode decomposition to identify linear and bilinear Koopman realizations from input–state data. Our open-loop analysis demonstrates that only the bilinear Koopman model can accurately capture nonlinear state–input couplings and quadratic terms essential for collision avoidance, whereas linear realizations fail to do so. We formulate a quadratic program for the robot path planning in the presence of moving obstacles in the lifted space and determine the optimal robot action in an MPC framework. Our approach is capable of finding the safe optimal action 320 times faster than a nonlinear MPC counterpart that solves the path planning problem in the original state space. Our work highlights the potential of bilinear Koopman realizations for linearization of highly nonlinear optimal control problems subject to nonlinear state and input constraints to achieve computational efficiency similar to linear problems.

\end{abstract}

%%%%%%%%%%%%%%%%%%%%%%%%%%%%%%%%%%%%%%%%%%%%%%%%%%%%%%%%%%%%%%%%%%%%%%%%%%%%%%%%
\section{INTRODUCTION}

Safe and efficient navigation of autonomous robots in dynamic environments is a long-standing challenge in robotics and control. Robots must continuously plan trajectories that avoid both static and moving obstacles while making progress toward mission objectives. This problem combines nonlinear dynamics with inherently nonconvex collision avoidance constraints \cite{park2024survey}. Optimization-based motion planning and control have achieved strong results, yet directly handling these nonlinear constraints within receding horizon frameworks often leads to significant computational burden that limits real-time applicability on embedded platforms \cite{lutz2021efficient}. Recent work has explored convex formulations and linear Model Predictive Control (MPC) variants to improve tractability \cite{najafqolian2024control}, nonetheless, maintaining safety and accuracy in highly dynamic scenes remains difficult. These challenges motivate the development of models and controllers that fully realize the nonlinear structure of the problem to ensure safety while enabling efficient convex optimization at runtime.

Nonlinear MPC can directly enforce nonlinear dynamics and safety-critical constraints but suffers from high computational cost for real-time use \cite{allamaa2024real}. Similar trade-offs between accuracy and tractability have been noted \cite{abdolmohammadi2025optimal}. To address this, convex and linearized MPC formulations have been explored, such as successive convexification \cite{uzun2024successive} and convex quadratic programs \cite{rakovic2021convex}. Control Barrier Functions (CBF) have also been integrated with MPC to encode obstacle avoidance efficiently, including chance-constrained MPC with CBFs \cite{li2023moving}, dynamic CBF-MPC for mobile robots \cite{jian2022dynamic}, and adaptive dynamic CBF-MPC for unmanned Ground vehicles (UGV) in unstructured terrain \cite{guo2024obstacle}.

Data-driven modeling has emerged as an alternative to analytic linearization for motion planning and control. Recently, Koopman operator theory has gained attention as it provides a linear representation of nonlinear dynamics in a lifted observable space, enabling prediction and control with linear tools \cite{korda2018linear}. Practical approximations include Dynamic Mode Decomposition (DMD) and Extended  Dynamic Mode Decomposition (EDMD) with control, which identify finite-dimensional predictors directly from input--state data \cite{proctor2016dynamic}. These methods have been successfully applied to vehicle dynamics and MPC design \cite{cibulka2020model}, driver--automation shared control \cite{guo2023koopman}, and robot navigation under uncertainty \cite{soltanian2025pace,zhang2025koopman}. More recent works also integrate Koopman models with collision avoidance through control barrier functions \cite{folkestad2020data} or collision-aware planning with learned Koopman dynamics \cite{chen2025dk}. Deep Koopman structures have also been introduced, where neural networks are used to jointly learn the lifting functions and linear predictors, improving accuracy and scalability for complex vehicle dynamics \cite{abtahi2025multi,xiao2022deep}. While these approaches demonstrate the potential of Koopman models for safety-critical navigation, they often rely on auxiliary safety layers or are integrated within sampling-based planning frameworks. As a result, the collision-avoidance constraints are not represented directly in the lifted linear model, which can limit their efficiency or generalizability in real-time MPC settings.

In this work, we develop a data-driven framework for motion planning and control of mobile robots that overcomes two major bottlenecks of existing approaches: the nonlinear dynamics of the mobile robot and the nonlinear nature of collision-avoidance constraints. While NMPC can handle both, its computational cost makes real-time operation infeasible in highly interactive environments. Linear Koopman MPC, on the other hand, is tractable but cannot faithfully capture the nonlinear behaviors required for safe navigation. However, bilinear Koopman realizations provide a balanced alternative, preserving linearity in the lifted space while capturing essential input–state couplings that enhance modeling accuracy \cite{bruder2021advantages, abtahi2025deep}.

Our approach bridges this gap by identifying a bilinear Koopman realization of the unicycle dynamics using bilinear EDMD. The key insight is that the same bilinear lifting also generates the essential nonlinear terms that describe collision-avoidance constraints. By propagating these terms in the lifted space, both the robot dynamics and the collision-avoidance inequalities admit a linear representation. This allows the entire planning and control problem to be formulated as a convex Quadratic Programming Model Predictive Controller (QP-MPC) that maintains the fidelity of NMPC while retaining the computational efficiency of linear methods. We further demonstrate that the identified bilinear model precisely captures the input–state couplings responsible for obstacle interactions, as revealed by the learned high-dimensional matrices.

The main contributions of this paper are:
\begin{itemize}
    \item \textbf{EDMD-based bilinear Koopman modeling.} A bilinear Koopman realization is identified that captures both unicycle dynamics and the nonlinear terms of collision-avoidance constraints.
    \item \textbf{Validation of bilinear couplings.} Analysis shows that the learned coupling matrices accurately represent key input--state interactions for safe planning.
    \item \textbf{Convex QP-MPC formulation.} The lifted linear structure enables efficient QP-MPC for real-time navigation in dynamic environments.    
\end{itemize}

The rest of this paper is organized as follows. Section~\ref{sec:dynamics} introduces the unicycle model and collision-avoidance constraints. Section~\ref{sec:koopman} presents the linear and bilinear Koopman realizations via EDMD, followed by a comparison of their modeling accuracy. Section~\ref{sec:mpc} formulates the proposed QP-MPC with lifted dynamics and constraints. Section~\ref{sec:results} reports simulation studies, including a performance comparison with NMPC, showing that the proposed controller achieves comparable accuracy with much lower computation time. Section~\ref{sec:conclusion} concludes the paper.

%%%%%%%%%%%%%%%%%%%%%%%%%%%%%%%%%%%%%%%%%%%%%%%%%%%%%%%%%%%%%%%%%%
%%%%%                       dynamics                  %%%%%%%%%%%%
%%%%%%%%%%%%%%%%%%%%%%%%%%%%%%%%%%%%%%%%%%%%%%%%%%%%%%%%%%%%%%%%%%
\section{Dynamics and Constraints}\label{sec:dynamics}
\subsection{Mobile Robot Modeling}\label{subsec:unicycle}
The unicycle kinematic model is widely used in robotics since it provides a simple yet effective representation of differential-drive platforms and wheeled ground vehicles. In this work, the mobile robot is described by this model, with the state vector defined as
\begin{equation}
    x = \begin{bmatrix} X & Y & v & \theta \end{bmatrix}^\top \in \mathbb{R}^{n=4},
\end{equation}
where $X, Y$ denote the global position of robot, $v$ is the forward velocity in local frame, and $\theta$ is the heading angle. The control input vector is
\begin{equation}
    u = \begin{bmatrix} a & \omega \end{bmatrix}^\top \in \mathbb{R}^{m=2},
\end{equation}
where $a$ denotes the longitudinal acceleration and $\omega$ the angular velocity input. A schematic representation of the unicycle model used in this work is shown in Fig.~\ref{fig:obstacle_robot}, and the continuous-time kinematics are expressed as
\begin{equation}
    \dot{x} =
    \begin{bmatrix}
        v \cos\theta \\
        v \sin\theta \\
        a \\
        \omega
    \end{bmatrix},
    \label{eq:unicycle_dynamics}
\end{equation}
which capture the evolution of position and orientation under forward velocity and heading changes. 
For data generation and control implementation, \eqref{eq:unicycle_dynamics} is discretized using a fourth-order Runge--Kutta method with a zero-order hold on inputs and a fixed sampling period of $t_s = 0.1\,\text{s}$.

\subsection{Nonlinear Collision-Avoidance Constraints}\label{subsec:collision}
\begin{figure}[t]
      \centering
      
      \includegraphics[width=\linewidth]{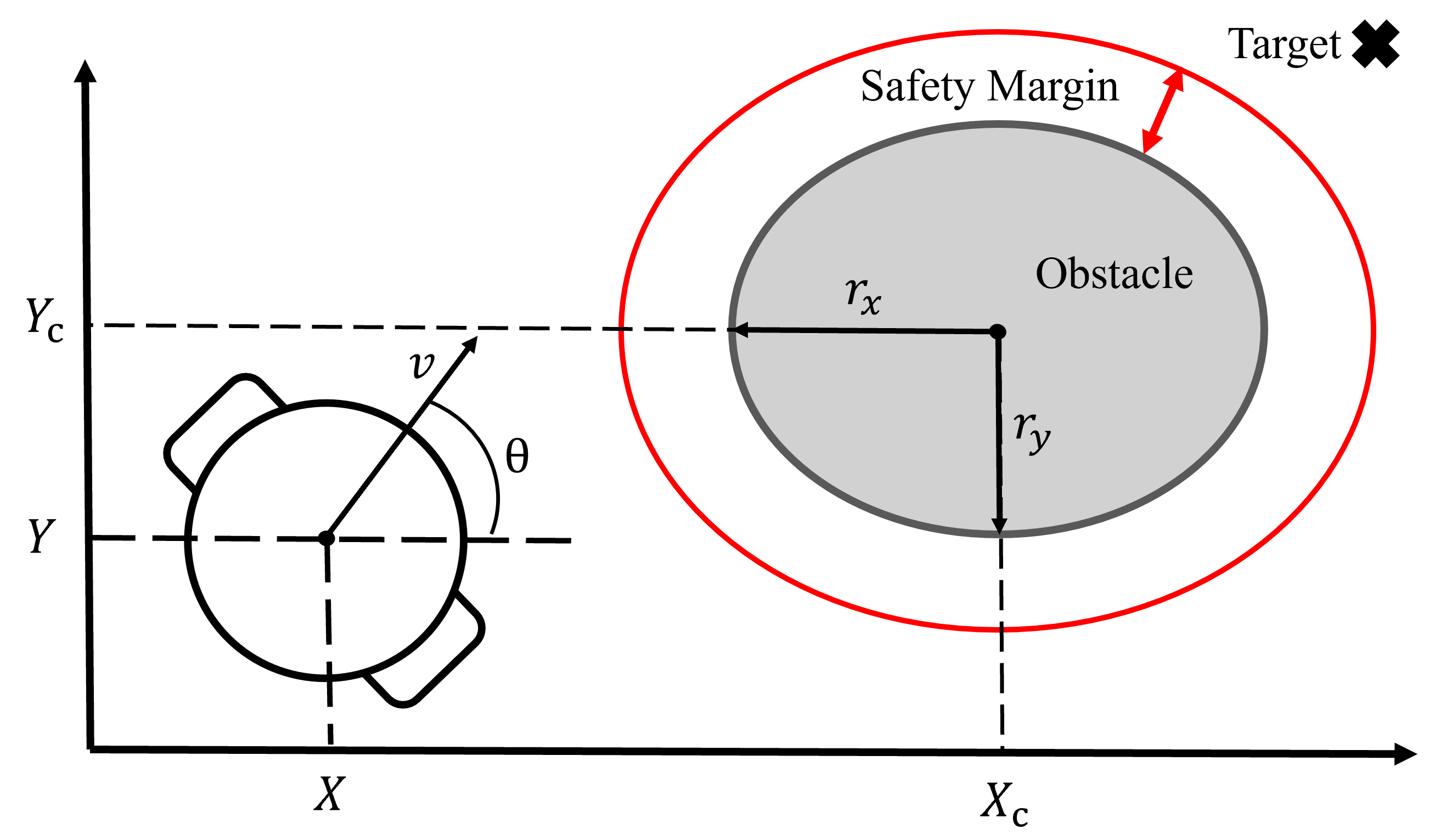}
      \caption{Unicycle robot kinematics and ellipsoidal obstacle representation with safety margin for collision avoidance.}
      \label{fig:obstacle_robot}
\end{figure}

In highly interactive robotic environments, obstacles can be represented in various forms, such as rectangular or polygonal bounding boxes, circular approximations, or velocity obstacles for dynamic agents \cite{haraldsen2023dynamic,han2022reinforcement}. Ellipsoidal models are particularly attractive because they provide smooth boundaries that approximate diverse obstacle shapes while yielding analytic quadratic inequalities suitable for optimization-based control \cite{rosenfelder2025efficient,brito2019model}. Motivated by these advantages, we represent obstacles as ellipsoidal keep-out regions in the plane, as illustrated in Fig.~\ref{fig:obstacle_robot}. At prediction step $k$, an obstacle with center $(X_{c,k}, Y_{c,k})$ and semi-axes $r_x$ and $r_y$ defines the admissible robot positions as
\begin{equation}
    \frac{(X_k - X_{c,k})^{2}}{r_x^{2}}
    + \frac{(Y_k - Y_{c,k})^{2}}{r_y^{2}}
    \ge 1 + \epsilon,
    \label{eq:collision_const}
\end{equation}
where $\epsilon > 0$ is a prescribed safety margin. This inequality ensures that the robot’s position $(X_k, Y_k)$ remains outside the obstacle with buffer.

%%%%%%%%%%%%%%%%%%%%%%%%%%%%%%%%%%%%%%%%%%%%%%%%%%%%%%%%%%%%%%%%%%
%%%%%                       Koopman                   %%%%%%%%%%%%
%%%%%%%%%%%%%%%%%%%%%%%%%%%%%%%%%%%%%%%%%%%%%%%%%%%%%%%%%%%%%%%%%%

\section{Koopman Operator Framework}\label{sec:koopman}

\subsection{Koopman Operator}
The Koopman operator, introduced in dynamical systems theory \cite{koopman1931hamiltonian}, provides a linear infinite-dimensional representation of nonlinear dynamics by acting on a space of observable functions of the state. By lifting the system states into this higher-dimensional space, a linear operator equivalently describes the nonlinear evolution of the original system. Direct computation with the infinite-dimensional Koopman operator is generally intractable. However, for a given finite set of observables, a finite-dimensional linear operator can be approximated that acts within the subspace spanned by the observable set. 

For a discrete-time nonlinear system with state $x_k \in \mathbb{R}^n$, the dynamics can be lifted into a higher-dimensional embedding space through a set of vector-valued observables defined as  
\begin{equation}
    Z_k = [\psi_1(x_k),\, \psi_2(x_k),\, \ldots,\, \psi_{N_\ell}(x_k)]^\top,
\end{equation} 
where each $\psi_i:\mathbb{R}^n \to \mathbb{R}$ is a scalar-valued observable function, and $N_\ell \in \mathbb{N}$ denotes the total number of selected observables. Within this subspace, the infinite-dimensional Koopman operator can be approximated by a finite-dimensional matrix $\mathbf{K}$ such that  
\begin{equation}
    Z_{k+1} = \mathbf{K} Z_k.
    \label{eq:koopmanmatrix}
\end{equation}
Such a formulation enables the evolution of nonlinear system measurements to be represented as a linear state-transition in the lifted space, providing a foundation for prediction and control using linear techniques.

\subsection{ Linear EDMD with Control }

EDMD is a widely used data-driven method for approximating the Koopman operator by projecting it onto the span of a chosen set of observables. For controlled systems, the lifted subspace must also account for control inputs. Therefore, at each snapshot, the observable set $Z_k$ needs to be augmented with the corresponding input $u_k$ to form the regression data. For a dataset of snapshot pairs, we define
\begin{align}
    \Phi   &=[({Z_0}^\top,u_0^\top)^\top,({Z_1}^\top,u_1^\top)^\top, ..., ({Z_P}^\top,u_P^\top)^\top]\\
    \Phi^+ &=[({Z_1}^\top,u_1^\top)^\top,({Z_2}^\top,u_2^\top)^\top, ..., ({Z_{P+1}}^\top,u_{P+1}^\top)^\top]\notag
\end{align}
where $P$ denotes the total number of snapshots. A finite-dimensional Koopman matrix is then identified by solving a least-squares regression problem
\begin{equation}
    \min_{\textbf{K}}  \left\| \Phi^+ - \textbf{K}\Phi \right\|_F.
    \label{eq:regresion}
\end{equation}
which admits the closed-form solution $\mathbf{K} = \Phi^+ \Phi^{\dagger}$ using the Moore--Penrose pseudoinverse and takes the structured form
\begin{equation}
   \textbf{K} = \begin{bmatrix}
A_{\{N_l \times N_l \}} & B_{\{N_l \times m \}} \\
\cdot & \cdot
\end{bmatrix}.
    \label{eq:koopman}
\end{equation}
Then, the system dynamics under the approximated Koopman operator can be expressed as
\begin{equation}
\begin{bmatrix}
Z_{k+1} \\
u_{k+1}
\end{bmatrix}
=
\begin{bmatrix}
A & B \\
\cdot & \cdot
\end{bmatrix}
\begin{bmatrix}
Z_{k} \\
u_{k}
\end{bmatrix}.
\end{equation}
Note that the control inputs are commanded externally and do not follow autonomous dynamics \cite{Proctor2016}. Therefore, identification focuses only on the top block, yielding the reduced form
\begin{equation}
Z_{k+1} = A Z_{k} + B u_k.
\label{eq:koopman_linear}
\end{equation}

\subsection{Bilinear EDMD with Control}\label{subsec:bilinear_edmd}
Linear Koopman realizations consider only linear input–state relations, which limit their ability to capture multiplicative couplings. In contrast, bilinear realizations preserve linearity in the lifted state while admitting multiplicative interactions with the inputs. Importantly, it has been shown that every control-affine nonlinear system admits an infinite-dimensional bilinear realization, whereas a valid linear realization is not guaranteed to exist \cite{bruder2021advantages}. Consequently, bilinear Koopman models are often more expressive and better suited for approximating nonlinear systems from data. Since the unicycle dynamics in \eqref{eq:unicycle_dynamics} are control-affine, there exists a bilinear Koopman realization capable of representing these dynamics in a higher-dimensional lifted space.

To identify a finite-dimensional bilinear predictor with EDMD, the lifted representation is augmented with both the control inputs and their products with the lifted state, constructing the snapshot pairs as follows
\begin{align}
    \Phi   &=[({Z_0}^\top,u_0^\top, (u_0 \otimes Z_0)^\top)^\top, ... \nonumber\\
           &\quad\quad\quad\quad,({Z_P}^\top,u_P^\top,(u_P \otimes Z_P)^\top)^\top] \\
    \Phi^+ &=[({Z_1}^\top,u_1^\top, (u_1 \otimes Z_1)^\top)^\top, ... \nonumber\\
           &\quad\quad\quad\quad, ({Z_{P+1}}^\top,u_{P+1}^\top,(u_{P+1} \otimes Z_{P+1})^\top)^\top],
\end{align}
where the input–state interaction terms are constructed using the Kronecker product $u_k\otimes Z_k\in\mathbb{R}^{mN_\ell}$.
By solving the least-squares regression problem in \eqref{eq:regresion}, the following bilinear Koopman realization is obtained
\begin{equation}
       \begin{bmatrix}
           Z_{k+1}\\u_{k+1}\\u_{k+1} \otimes Z_{k+1}
       \end{bmatrix} =
    \begin{bmatrix}
        A & B & H_1 &\cdots &H_m \\
        \vdots & \vdots & \vdots & \ddots & \vdots
    \end{bmatrix}
    \begin{bmatrix}
           Z_{k}\\u_{k}\\u_{k} \otimes Z_{k}
       \end{bmatrix}.
    \label{eq:lifted_dynamics}
\end{equation}
Focusing solely on the evolution of the lifted observables, the Koopman approximation is expressed as
\begin{equation}
    Z_{k+1} = AZ_{k} + B u_k + \sum_{i=1}^m H_i \left[u^i_{k} \cdot  Z_k \right],
    \label{eq:bilinear}
\end{equation}

where $H_i \in \mathbb{R}^{N_\ell \times N_\ell}$ captures the bilinear coupling between the lifted state and the $i$-th input channel. This structure reduces to the linear EDMD realization when all input–state coupling matrices vanish, that is, $H_i = 0$.

\subsection{Koopman Models for Path Planing of Mobile Robot}\label{subsec:koopman_modeling}

A key step in constructing Koopman-based models is the choice of observables. The goal is to define a lifted representation that captures the nonlinear dynamics of the unicycle model while also embedding the nonlinear collision-avoidance constraints into a linear form suitable for predictive control. The lifting process begins with a choice of preset dictionary $\hat\Psi(x)$ designed to encode the essential nonlinearities of the system. For the unicycle dynamics \eqref{eq:unicycle_dynamics}, trigonometric functions of the heading angle and mixed velocity–angle terms are included to represent the nonlinear motion. For the collision-avoidance constraint \eqref{eq:collision_const}, quadratic terms in the position states, $X^2$ and $Y^2$, are explicitly introduced. This choice is motivated by the fact that the nonlinear quadratic inequality governing obstacle avoidance becomes affine in the lifted observables, as will be discussed in detail in Section~\ref {subsec:QP_constraints}.
Accordingly, the preset dictionary is defined as
\begin{align}
    \hat\Psi(x) &= [\hat\psi_1, \hat\psi_2, \ldots, \hat\psi_{10}]^\top \nonumber\\
    &= \big[\, X,\; Y,\; v,\; \theta,\; X^2,\; Y^2, \nonumber\\
    &\quad\;\; \sin\theta,\; \cos\theta,\; v\sin\theta,\; v\cos\theta \,\big]^\top,
    \label{eq:psi_base}
\end{align}
and the lifted dictionary $Z$ is then constructed by including all polynomial combinations of $\hat\Psi(x)$ up to degree $\rho = 2$, which is sufficient to capture the dominant nonlinearities in the dynamics, i.e.,
\begin{align}
     Z &= \big[\psi_1,\; \psi_2,\; \ldots,\; \psi_{N_l}\big]^\top \in \mathbb{R}^{N_l},
    \label{eq:Z_lifting}\\
    \text{s.t.} \quad 
    \psi_j(\hat\Psi) &\in 
    \left\{ \prod \hat\psi_i^{\rho_i} \;\middle|\; \sum \rho_i \leq \rho,\; \forall \rho_i \geq 0 \right\},
    \label{eq:psi2} 
\end{align}
which result in $N_\ell = 65$ for the chosen setup.
This finite dictionary provides the common lifted subspace on which both linear and bilinear Koopman realizations are identified using EDMD, allowing a direct comparison of their predictive capabilities.

The dataset for model identification consisted of $100{,}000$ trajectories, each corresponding to four seconds. For every trajectory, a random control sequence of acceleration and angular velocity inputs was generated and applied to the discretized unicycle model with sampling time $t_s=0.1$. The dataset was divided into training and testing portions of $90\%$ and $10\%$, respectively, yielding a total of $P=3.6$ million snapshot pairs for Koopman regression.

Each trajectory began at the origin $X=0, Y=0$ with an initial velocity sampled uniformly from the interval $[0,5]$\ \text{m/s} and a heading angle sampled uniformly from $[-\pi,\pi]$\ \text{rad}. Fixing the initial position at the origin stabilized the dataset by ensuring bounded planar coordinates over the four-second horizon. This approach requires the MPC formulation to shift coordinates into the local robot frame at each step in the next steps. 
\begin{figure}[t]
      \centering
      
      \includegraphics[width=\linewidth]{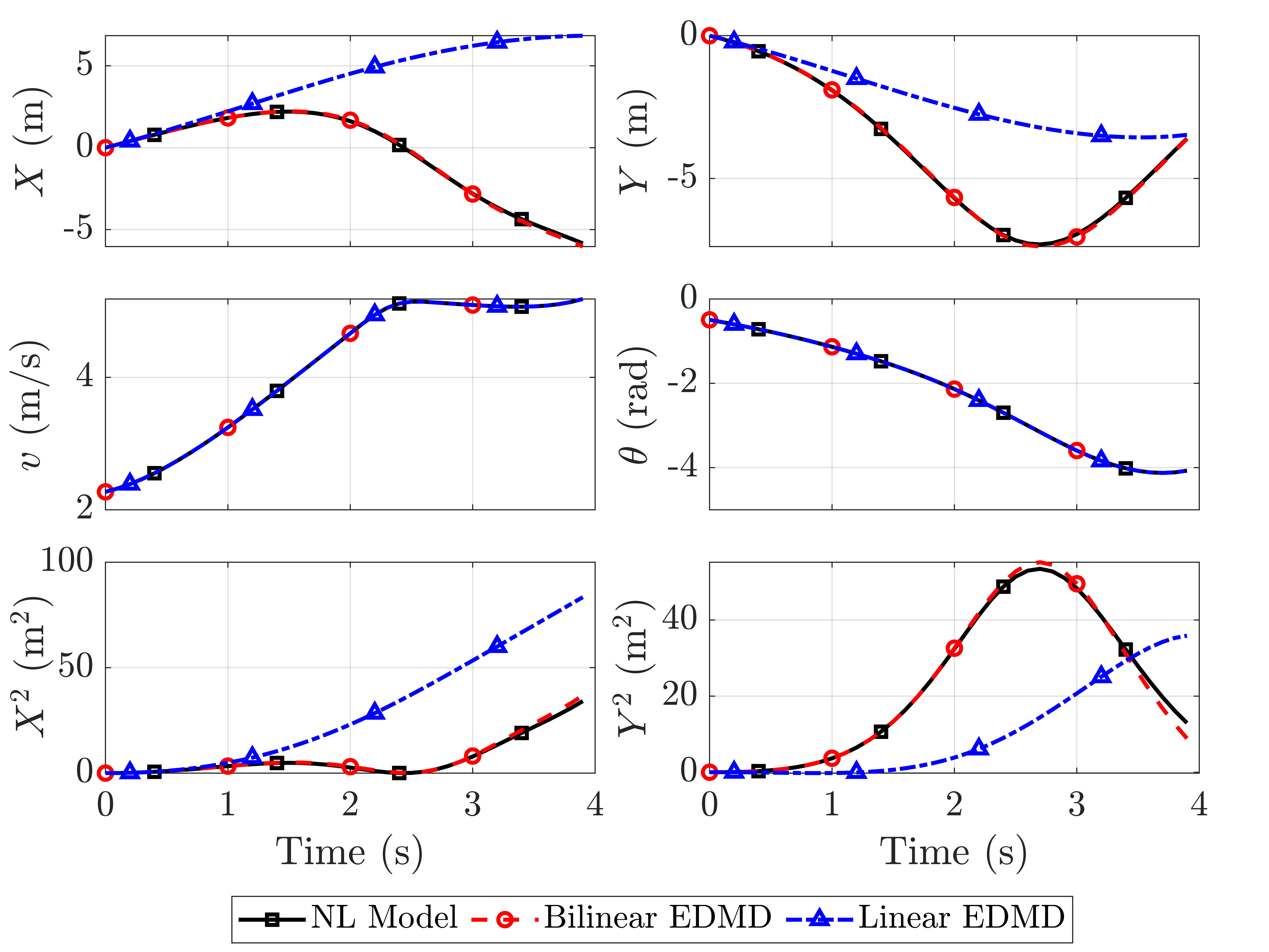}
      \caption{Open-loop prediction of the nonlinear unicycle model (black) compared with Bilinear EDMD (red) and Linear EDMD (blue) over a 4-second horizon. The bilinear realization accurately reproduces both the primary states $(X,Y,v,\theta)$ and the quadratic observables $(X^2,Y^2)$.}
      \label{fig:openLoop_bi vs lin}
\end{figure}
\begin{table}[t]
    \centering
    \setlength{\tabcolsep}{6pt}
    \caption{Average open-loop prediction RMSE over all $10k$ test trajectories for Linear EDMD vs. Bilinear EDMD.}
    \label{tab:rmse_edmd_bilinear}
    \begin{tabular}{lcc}
        \toprule
        \textbf{State Variable} & \textbf{Linear EDMD} & \textbf{Bilinear EDMD} \\
        \midrule
        $X$ (m)            & 3.118  & \textbf{0.116} \\
        $Y$ (m)            & 3.257  & \textbf{0.117} \\
        $v$ (m/s)          & \textbf{$9.2\mathrm{e}{-8}$} & $9.4\mathrm{e}{-8}$ \\
        $\theta$ (rad)     & \textbf{$6.5\mathrm{e}{-8}$} & $6.8\mathrm{e}{-8}$ \\
        $X^{2}$ (m$^{2}$)  & 31.682 & \textbf{2.070} \\
        $Y^{2}$ (m$^{2}$)  & 30.087 & \textbf{2.087} \\
        \bottomrule
    \end{tabular}
\end{table}
Figure~\ref{fig:openLoop_bi vs lin} compares the open-loop predictions of identified linear EDMD and bilinear EDMD models against the nonlinear unicycle ground truth for a representative test input sequence, while Table~\ref{tab:rmse_edmd_bilinear} summarizes the average RMSE across all test trajectories. Both realizations reproduce the velocity and heading states with near-perfect accuracy, consistent with the unicycle dynamics in \eqref{eq:unicycle_dynamics}, where $v$ and $\theta$ depend linearly on the control inputs $a_k$ and $\omega_k$. However, the bilinear model provides substantially higher accuracy for the integrated position states $X$ and $Y$, as well as the quadratic observables $X^2$ and $Y^2$. In contrast, the linear realization systematically underestimates these effects, leading to significant deviations.  This distinction arises from the discrete-time dynamics. For example, the evolution of $X$ and $X^2$ can be approximated using a second-order Taylor expansion as
\begin{align}
    X_{k+1} &\approx X_k + v_k \cos(\theta_k) t_s \notag \\
    &\quad + \tfrac{1}{2} t_s^2 \Big(
    \underbrace{a_k \cos(\theta_k)}
    - \underbrace{v_k \omega_k \sin(\theta_k)}\Big),
    \label{eq:taylor1} \\
    &\qquad\qquad\quad u_k^1Z_k^7\qquad\quad\ \ u_k^2Z_k^8 \notag \\
    X_{k+1}^{2} &\approx X_k^{2} + 2X_k v_k \cos(\theta_k) t_s \notag \\
    &\quad + t_s^{2} \Big(
    \underbrace{X_k a_k \cos(\theta_k)}
    - \underbrace{X_k v_k \omega_k \sin(\theta_k)} \Big) ,
    \label{eq:taylor2}\\
    &\qquad\qquad\quad u_k^1Z_k^{27}\qquad\quad\ \ u_k^2Z_k^{28} \notag
\end{align}
where $a_k$ and $\omega_k$ are denote as $u_k^1$ and $u_k^2$, respectively, and $Z_k^i$ denotes the $i$-th observable in the dictionary. The underbraced terms highlight the multiplicative input–state couplings that cannot be represented by a purely linear Koopman realization but are naturally incorporated in the bilinear model. The same structure appears in the $Y$ and $Y^2$ expansions.

\begin{figure}[thpb]
      \centering
      
      \includegraphics[width=\linewidth]{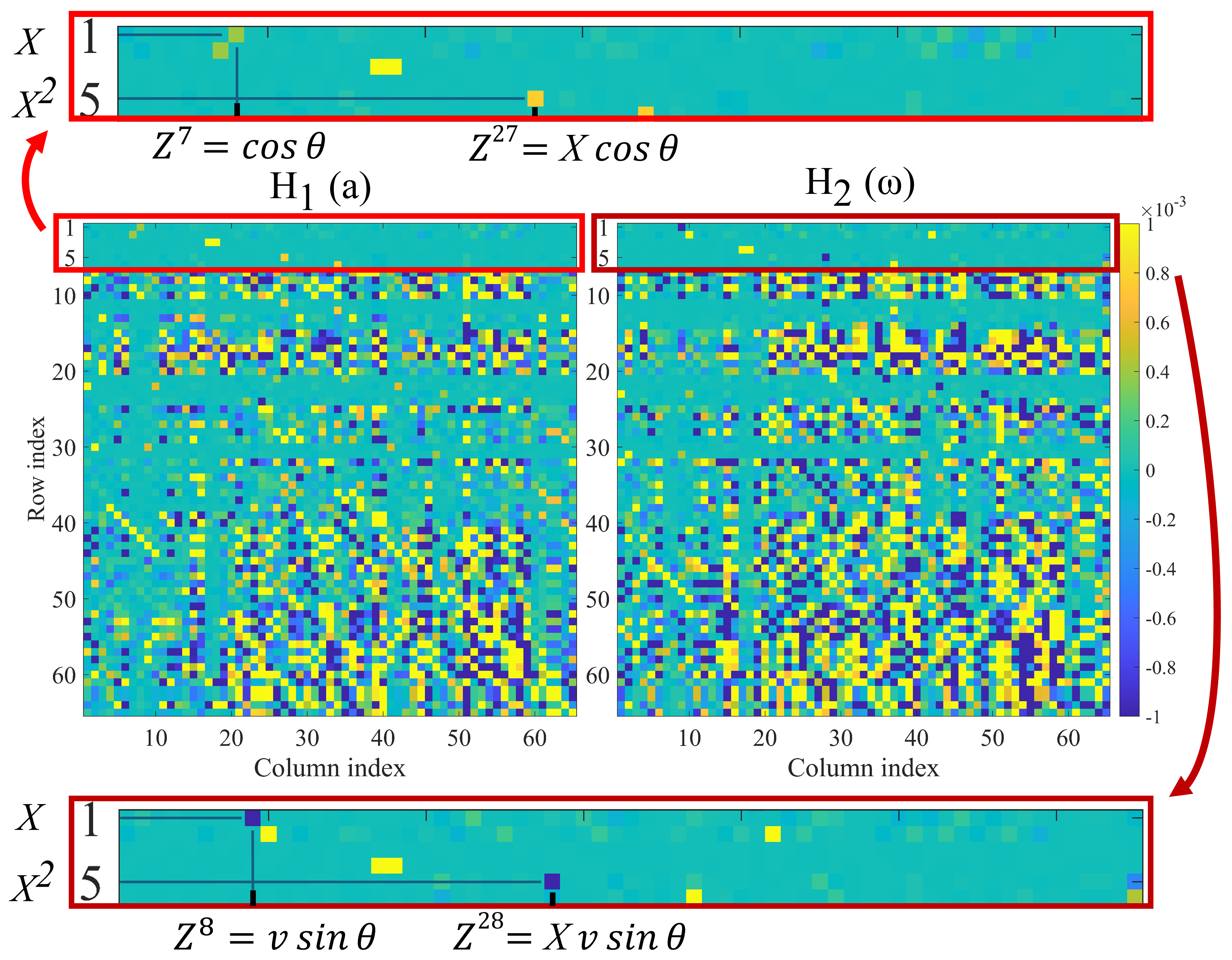}
      \caption{Identified bilinear coupling matrices $H_1$ (acceleration channel) and $H_2$ (angular velocity channel). The highlighted entries show the learned input–state interactions responsible for the bilinear terms in \eqref{eq:taylor1} and \eqref{eq:taylor2}.}
      \label{fig:H1_H2}
\end{figure}

Figure~\ref{fig:H1_H2} shows the coupling matrices $H_1$ and $H_2$ for the identified bilinear model. For the state $X$, the bilinear terms in \eqref{eq:taylor1} correspond to the interaction of input $u^1$ (acceleration) with observable $Z^7=\cos\theta$, and of input $u^2$ (angular velocity) with observable $Z^8=v\sin\theta$. These interactions are encoded in $H_1$ and $H_2$, respectively, and appear as structured patterns in the corresponding rows of the coupling matrices. Similarly, in \eqref{eq:taylor2}, the dynamics of $X^2$ involve couplings between $u^1$ and $Z^{27}=X\cos\theta$, and between $u^2$ and $Z^{28}=Xv\sin\theta$, which are again captured by the bilinear matrices. The presence of these learned couplings in $H_1$ and $H_2$ confirms that the bilinear realization accurately identifies and encodes input–state multiplicative interactions directly from data.

%%%%%%%%%%%%%%%%%%%%%%%%%%%%%%%%%%%%%%%%%%%%%%%%%%%%%%%%%%%%%%%%%%
%%%%%                       BK_MPC                    %%%%%%%%%%%%
%%%%%%%%%%%%%%%%%%%%%%%%%%%%%%%%%%%%%%%%%%%%%%%%%%%%%%%%%%%%%%%%%%

\section{Bilienar Koopman MPC}\label{sec:mpc}

Accurate position control is essential when a mobile robot navigates in interactive environments with moving obstacles. Using the identified bilinear Koopman predictor, the planning problem is expressed in a lifted space where both the dynamics and the collision-avoidance constraints admit a linear structure. At each sampling instant, the bilinear term is linearized at the initial lifted state, which yields a linear time-varying prediction model and enables a convex quadratic program (QP) for MPC at each iteration.

\subsection{Koopman QP-MPC}

We formulate a QP-MPC based on the bilinear Koopman model identified. The general QP formulation for minimizing $J$ with respect to the stacked input vector $\mathbf{U}$ can be written as
\begin{equation}
    J = \tfrac{1}{2} \textbf{U}^\top \textbf{H} \textbf{U} + \textbf{f}^\top \textbf{U} ,\quad \text{s.t.} \quad \textbf{G}\textbf{U}\le \textbf{D}
    \label{eq: QP cost and constraint}
\end{equation}
where the linear inequalities define the feasible region. Within this region, the convex quadratic program minimizes $J$ according to the weighting matrices $\mathbf{H}$ and $\mathbf{f}$. In the MPC setting, the cost function typically consists of two terms: a tracking error penalty, which enforces progress towards the target, and a control effort penalty, which ensures efficient and smooth motion. The MPC cost function for the mobile robot is defined as
\begin{equation}
    J^{mr}_{1\to N_h} = \sum_{k=1}^{N_h} \| y_k - r_{tg}\|_Q^2 + \sum_{k=0}^{{N_h}-1} \|u_k\|_R^2,
\end{equation}
where $N_h$ is the prediction horizon. The first term penalizes the tracking error between the system output $y_k=[X_k,\,Y_k,\,v_k,\,\theta_k]^\top$ and the target point $r_{tg}=[X_{tg},\,Y_{tg},\,v_{tg},\,\theta_{tg}]^\top$ while the second term penalizes the control effort.
The output vector $y_k$ is obtained from the predicted lifted state $\hat{Z}_k$ through a projection matrix $C$,
\begin{equation}
y_k = C \hat{Z}_k, \qquad
C = [I_{4\times 4}, 0_{4\times N_\ell}],
\end{equation}
where $I_{4\times 4}$ extracts the physical robot states and $N_\ell$ is the total dimension of the lifted space.

The weighting matrices $Q$ and $R$ trade off target tracking versus control effort. To enable a QP-MPC formulation, the control inputs over the prediction horizon are stacked as
\begin{equation}
\mathbf{U} = [u_0^\top, u_1^\top, \dots, u_{N_h-1}^\top]^\top.
\label{eq:stackU}
\end{equation}
Given the initial mobile robot state $x_0$, the global coordinate frame is first shifted to the robot’s local frame so that planar positions start at zero, consistent with the training data. In this shifted frame, obstacle and target positions are also redefined accordingly. The shifted state is then mapped with the preset dictionary $\hat\Psi_0$ as in \eqref{eq:psi_base}. Finally, a polynomial lifting is applied to $\hat\Psi_0$ to obtain the full lifted state $Z_0$ as defined in \eqref{eq:Z_lifting}.

To obtain a convex formulation, the bilinear dependence is linearized at the current lifted state $Z_0$ at each prediction step. This yields a time-varying input matrix $B_t(Z_0)$, defined as
\begin{align}
    \hat{Z}_{k+1} &= A \hat{Z}_k + B_t(Z_0) u_k,\notag \\
    B_t(Z_0) &= B + \big[ H_1 Z_0 , H_2 Z_0 \big].
\end{align}
Using $Z_0$ and the Koopman matrices $A$ and $B_{t}$, the predicted future states evolve as
\begin{align}
\hat{Z}_1 &= A Z_0 + B_{t} u_0 \notag\\
\hat{Z}_2 &= A^{2} Z_0 + A B_t u_0 + B_t u_1 \notag\\
&\;\;\vdots \notag \\
\hat{Z}_{N_h} &= A^{N_h} Z_0 + A^{N_h-1} B_t u_0 + \cdots + B_t u_{N_h-1}.
\end{align}
The predicted lifted trajectory over the horizon can be written compactly as
\begin{equation}
    \hat{\textbf{Z}} = \boldsymbol{\Phi} Z_0 + \boldsymbol{\Gamma} \textbf{U},
\end{equation}
with coefficients
\begin{align}
    \hat{\textbf{Z}} &=\begin{bmatrix} \hat{Z}_1 \\ \hat{Z}_2 \\ \vdots \\ \hat{Z}_{N_h} \end{bmatrix}\quad
    \boldsymbol{\Phi} = \begin{bmatrix} A \\ A^2 \\ \vdots \\ A^{N_h} \end{bmatrix}, \notag\\
    \boldsymbol{\Gamma} &= \begin{bmatrix}
        B_{t} & 0 & \cdots & 0 \\
        A B_{t} & B_{t} & \cdots & 0 \\
        \vdots & \vdots & \ddots & \vdots \\
        A^{N_{h}-1} B_{t} & A^{N_{h}-2} B_{t} & \cdots & B_{t}
    \end{bmatrix}.
\end{align}

The predicted system outputs in dense form are then expressed as
\begin{equation}
    \textbf{Y} =  \textbf{C} \hat{\textbf{Z}}, 
\end{equation}
with $\textbf{C}= \mathrm{diag}_{N_h}(C,\dots,C)$.
Finally, the MPC cost $J^{\text{mr}}_{1\to N_h}$ can be expressed in dense form as
\begin{equation}
    J^{mr}_{1\to N_h} = \tfrac{1}{2} \textbf{U}^\top \textbf{H} \textbf{U} + \textbf{f}^\top \textbf{U},
\end{equation}
where
\begin{align}
    \textbf{H} &= 2\big(\boldsymbol{\Gamma}^\top \textbf{C}^\top \textbf{Q}\, \textbf{C} \boldsymbol{\Gamma} + \textbf{R}\big), \notag\\
    \textbf{f}^\top &= 2\Big((\textbf{C}\boldsymbol{\Phi} Z_0)^\top \textbf{Q}  -  \textbf{R}_{tg}^\top \textbf{Q}\Big)  \textbf{C}\boldsymbol{\Gamma},
\end{align}
with block-diagonal weight matrices 
\begin{align}
    \textbf{Q}&=\mathrm{diag}_{N_h}(Q,\dots,Q), \quad \textbf{R}=\mathrm{diag}_{N_h}(R,\dots,R),\notag \\[6pt]
    \textbf{R}_{tg}&=\mathbf{1}_{N_h} \otimes r_{tg}.
\end{align}
%\textbf{R}_{tg}&=[r_{tg}^\top,r_{tg}^\top,\cdots,r_{tg}^\top]^\top_{1\times4N_h}

\subsection{Linear Constraints and Collision Avoidance}
\label{subsec:QP_constraints}

\begin{figure*}[t]
  \centering
  \includegraphics[width=\textwidth]{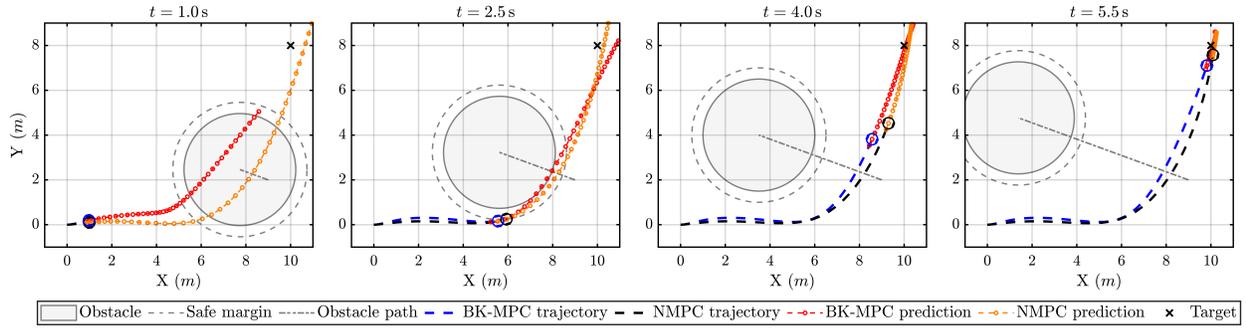}
  \caption{Closed-loop trajectories of BK-MPC and NMPC in the presence of a moving obstacle, shown at time snapshots $t=1.0\,\text{s}$, $t=2.5\,\text{s}$, $t=4.0\,\text{s}$, and $t=5.5\,\text{s}$. Predicted trajectories are indicated by dashed lines, while executed trajectories are shown as solid lines. Both controllers successfully reach the target while avoiding the moving obstacle. The proposed BK-MPC achieves real-time operation and is approximately $320$ times faster than NMPC, which does not meet real-time feasibility.}
  \label{fig:NMPC vs BKMPC}
\end{figure*}

The stacked input sequence \textbf{U} is subject to linear inequality constraints, which include input bounds and obstacle avoidance conditions. The input bounds are enforced as  

\begin{equation}
    G_u \mathbf{U} \le D_u, \quad
    G_u = \begin{bmatrix} I \\ -I \end{bmatrix}, \quad
    D_u = \begin{bmatrix} \mathbf{1}_{N_h} \otimes u_{\max} \\ -\mathbf{1}_{N_h} \otimes u_{\min} \end{bmatrix}.
\end{equation}
The collision-avoidance constraint for the moving ellipsoid object derived in \eqref{eq:collision_const} can be expressed linearly such that
\begin{equation}
    \alpha^1_k X_k + \alpha^2_k Y_k + \alpha^{5}_k X_k^2 + \alpha^{6}_k Y_k^2 \le \beta_k, \quad k=1,\dots,N_h,
    \label{eq:colision2}
\end{equation}
where
\begin{align}
    \alpha^1_k &= \frac{2X_{c,k}}{r_x^2}, \quad
    \alpha^2_k = \frac{2Y_{c,k}}{r_y^2}, \quad
    \alpha^{5}_k = -\frac{1}{r_x^2}, \quad
    \alpha^{6}_k = -\frac{1}{r_y^2}, \quad \notag\\
    \beta_k &= \frac{X_{c,k}^2}{r_x^2}+\frac{Y_{c,k}^2}{r_y^2} - 1 - \epsilon.
\end{align}
To capture the dynamics of the nonlinear terms appearing in \eqref{eq:colision2} and to propagate them independently of the state evolution, the measured output is extended to include the quadratic terms $X^2$ and $Y^2$. Such an output measurement from the predicted lifted state can be defined at each time step $k$ as 
\begin{equation}
    \tilde {y}_k=\tilde C \hat{Z}_k,\quad\quad \tilde C=[I_{6\times 6},0_{6\times N_l}]
    \label{eq:extended outputs}
\end{equation}
where $\tilde{y}_k \in \mathbb{R}^6$ contains the original states alongside with observables of $X^2$ and $Y^2$.
Using this extended output, the collision-avoidance constraint at each step can be written as
\begin{equation}
    s_k \tilde{y}_k<\beta_k,
\end{equation}
where
\begin{equation}
    s_k  = [\alpha^1_k , \alpha^2_k , 0 , 0 , \alpha^{5}_k,\alpha^{6}_k].
\end{equation}
Stacking over the horizon yields
\begin{equation}
    \textbf{S}\tilde{\textbf{Y}} <\boldsymbol{\beta},
    \label{eq:colisionstacked}
\end{equation}
where
\begin{align}
    \textbf{S} &=\mathrm{diag}(s_1,\ldots,s_{N_h}), \quad
     \boldsymbol{\beta}=[\beta_1, \beta_2, \cdots, \beta_{N_h}]^\top,\notag \\
     \tilde{\textbf{Y}} &= \tilde{\textbf{C}}  \textbf{Z},
\end{align}
with $\tilde{\textbf{C}} = \mathrm{diag}_{N_h}(\tilde C,\dots,\tilde C)$

Subsequently, the stacked constraint in \eqref{eq:colisionstacked} must be expressed in terms of the input vector \textbf{U} in order to align with the inequality constraints defined in \eqref{eq: QP cost and constraint}. This yields
\begin{equation}
     \textbf{G}_{obs}\textbf{U}  \le  \textbf{D}_{obs},
\end{equation}
where:
\begin{equation}
     \textbf{G}_{obs}=\textbf{S}\tilde{\textbf{C}} \boldsymbol{\Gamma}\\ \quad  \textbf{D}_{obs}= \boldsymbol{\beta} -\textbf{S} \tilde{\textbf{C}}\boldsymbol{\Phi} Z_0.
\end{equation}
The matrices $\mathbf{G}_{\mathrm{obs}}$ and $\mathbf{D}_{\mathrm{obs}}$ are inherently time-varying, as they depend on the obstacle trajectory assumed to be known or estimated, as well as the initial lifted state $Z_0$.
The QP to be solved at each sampling step is therefore
\begin{equation}
    \min_{\textbf{U}} \;\tfrac{1}{2} \textbf{U}^\top \textbf{H} \textbf{U} + \textbf{f}^\top \textbf{U}
    \quad \text{s.t.} \quad
    \begin{bmatrix}
        G_u \\  G_{\mathrm{obs}}
    \end{bmatrix} \mathbf{U}
    \le
    \begin{bmatrix}
        D_u \\ D_{\mathrm{obs}}
    \end{bmatrix}.
\end{equation}
After applying the first control input $u_0$, the lifted state is updated, the bilinear term is linearized at the new $Z_0$, and the optimization is repeated in a receding-horizon fashion.

%%%%%%%%%%%%%%%%%%%%%%%%%%%%%%%%%%%%%%%%%%%%%%%%%%%%%%%%%%%%%%%%%%
%%%%%                       results                   %%%%%%%%%%%%
%%%%%%%%%%%%%%%%%%%%%%%%%%%%%%%%%%%%%%%%%%%%%%%%%%%%%%%%%%%%%%%%%%

\section{Simulation and results}\label{sec:results}

The simulation study evaluates the proposed Bilinear Koopman MPC (BK-MPC) against a baseline nonlinear MPC (NMPC) under an interactive scenario characterized by the presence of a moving obstacle. The robot is initialized at the origin with velocity $v_0=0\,\text{m/s}$ and heading angle $\theta_0=0\,\text{rad}$. The target point state is defined as
$
r_{tg} = [\,10\ \text{m},\; 8\ \text{m},\; 0\ \text{m/s},\; 0\ \text{rad}\,]^\top 
$.
The admissible input set is restricted to
$
|a_k| \leq 2 \,\text{m/s}^2 $,$|\omega_k| \leq \pi \,\text{rad/s},
$
consistent with the bounds imposed during model identification. The NMPC baseline employs the exact nonlinear unicycle dynamics with nonlinear collision-avoidance constraints, while BK-MPC utilizes the bilinear Koopman model linearized at the current lifted state at each step. Both controllers share the same cost structure and weighting matrices. The sampling time is fixed at $t_s = 0.1\,\text{s}$ and the prediction horizon is $N_h=40$, corresponding to a four-second lookahead.
The stage cost matrices are
$
Q = \mathrm{diag}(1,\,1,\,0,\,0)$, $ R = \mathrm{diag}(4,\,10).
$
This design emphasizes terminal accuracy in position, while allowing velocity and orientation to adapt during obstacle interaction. Stronger input penalties enforce smooth control actions, with a higher weight on angular velocity to discourage abrupt maneuvers and promote feasible navigation.
For the example scenario, the obstacle is modeled as an ellipsoid with equal semi-axes $r_x = r_y = 2.5\ \text{m}$, and a safety margin of $\epsilon = 0.5$ is incorporated. The moving obstacle is initialized at $X_{c,0}=9\ \text{m}$ and $ Y_{c,0}= 4\ \text{m}$ and propagates with a constant velocity of $v_{\mathrm{obs}} = 1.5\ \text{m/s}$ along a heading of $\theta_{\mathrm{obs}} = 8\pi/9\ \text{rad}$.

\begin{figure}[t]
      \centering
      \includegraphics[width=\linewidth]{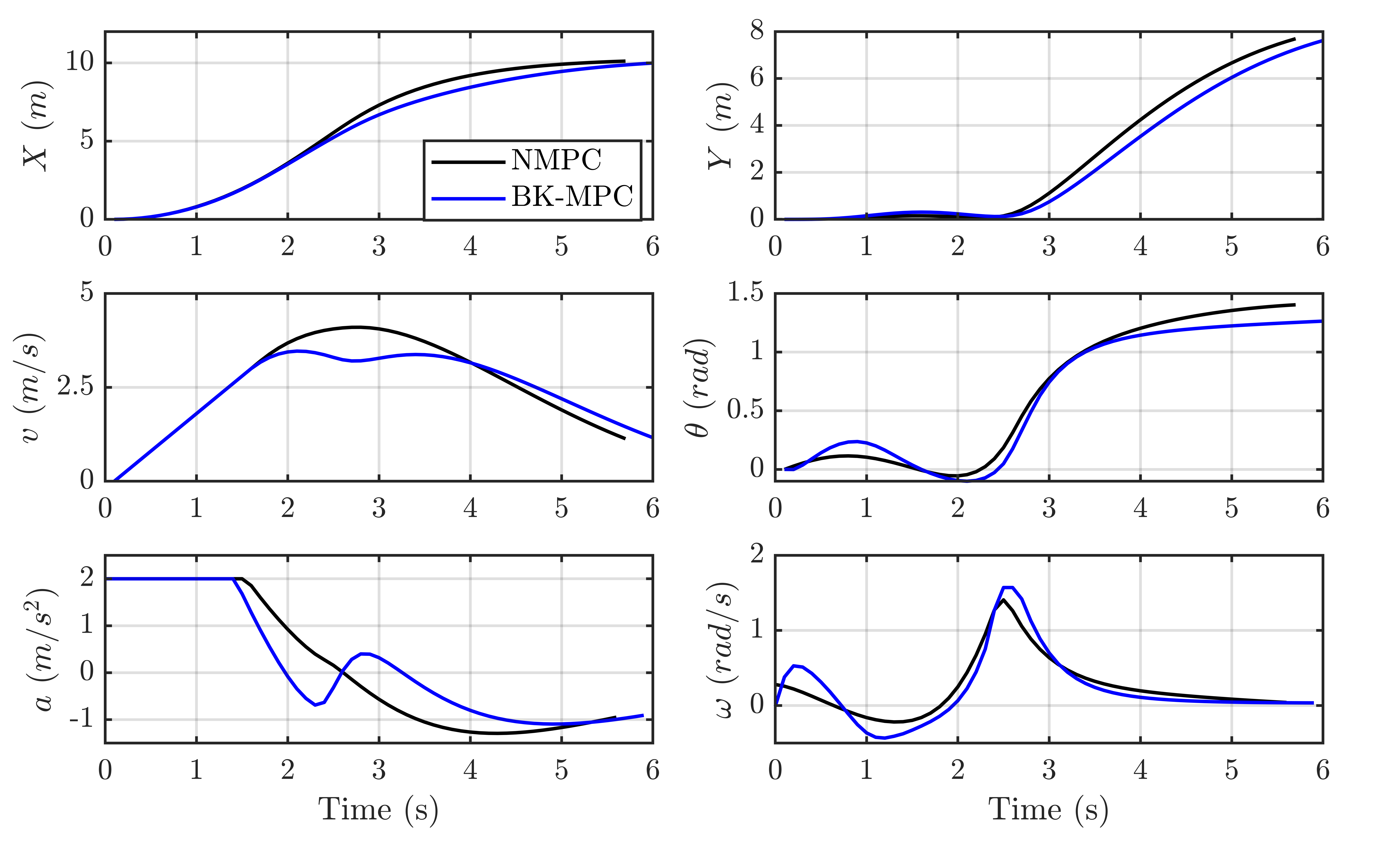}
      \caption{Closed-loop state and input trajectories of NMPC and BK-MPC. 
Both controllers exhibit nearly identical state trajectories and apply comparable control efforts while satisfying constraints.}
      \label{fig:State_input_traj}
\end{figure}

Figure~\ref{fig:NMPC vs BKMPC} illustrates the closed-loop trajectories of the mobile robot in the $X$--$Y$ plane under both NMPC and BK-MPC controllers. Both controllers successfully drive the robot to the target while respecting the safety constraints. At $t=2.5\,\text{s}$, the trajectories deviate to ensure obstacle avoidance with the prescribed safety margin $\epsilon$. Although the NMPC trajectory exhibits a slightly higher forward velocity around $t=4\,\text{s}$, both controllers guide the robot to the target with only a marginal difference in arrival time. Furthermore, Fig.~\ref{fig:State_input_traj} shows the closed-loop state and input trajectories of the robot for this interactive scenario. The executed trajectories by BK-MPC and NMPC are shown in blue and black, respectively. Both controllers produce nearly identical state evolutions and control inputs, indicating convergence to close optimal solutions for the given setup. These results demonstrate that the bilinear Koopman MPC recovers solutions comparable to NMPC, while its convex quadratic structure yields a significant reduction in computational burden compared to the nonconvex NMPC formulation.
   
\begin{figure}[t]
      \centering
      \includegraphics[width=\linewidth]{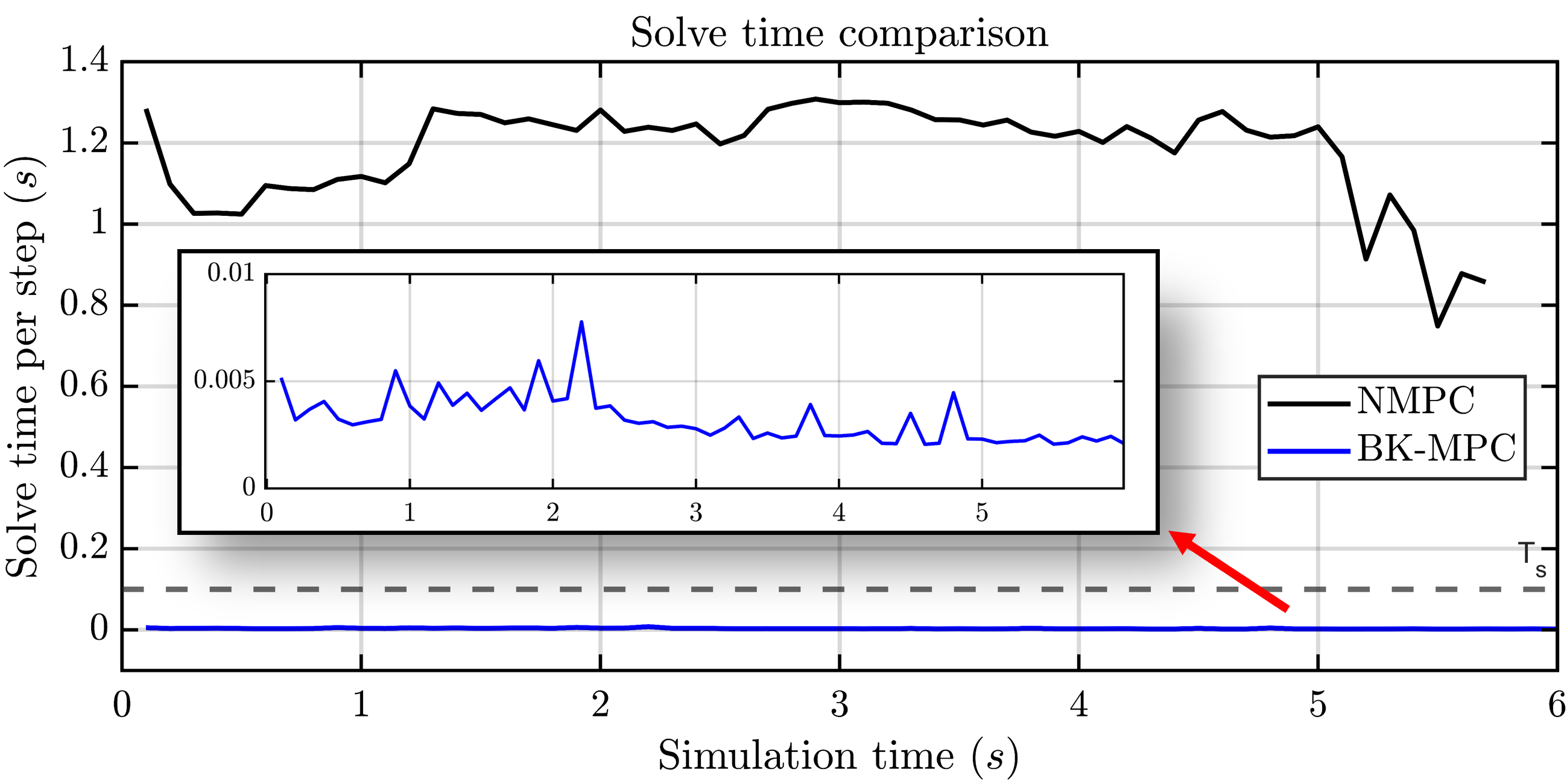}
      \caption{Per-step solve time trajectories of NMPC (black) and BK-MPC (blue) for the example scenario. The dashed line indicates the sampling time $t_s = 0.1\,\text{s}$.}
      \label{fig.solve time}
\end{figure}
A Monte Carlo study comprising 100 randomized target-following scenarios with different interactive moving obstacles was conducted to evaluate the computational performance of NMPC and BK-MPC. For each controller, the solve time per step was recorded across all runs, and the statistical metrics of average, maximum, and 95th-percentile values are summarized in Table~\ref{tab:solve_time}. The results demonstrate that the convex quadratic structure of BK-MPC yields approximately a $320$ times reduction in control computation time compared to the nonconvex NMPC formulation. Note that, all simulations were executed in \textsc{Matlab} R2024a, using the standard \texttt{quadprog} solver for BK-MPC and \texttt{fmincon} for NMPC, on a desktop machine equipped with an Intel\textsuperscript{\textregistered} Core\texttrademark\ i9-14900F CPU (24 cores, 2.0 GHz base frequency), 32 GB DDR5 RAM (5600 MT/s), and Windows 11 Home (64-bit).

Figure~\ref{fig.solve time} presents the per-step solve times of NMPC and BK-MPC for the example scenario discussed. The NMPC consistently exhibits solve times exceeding $0.7\,\text{s}$ per iteration, which is way larger than the sampling period $t_s = 0.1\,\text{s}$, thereby precluding real-time feasibility. In contrast, the BK-MPC achieves solve times below $0.008\,\text{s}$ across all iterations, remaining well within the sampling deadline. These results demonstrate that the convex bilinear Koopman formulation ensures real-time implementability, whereas the nonlinear NMPC formulation is computationally prohibitive for online operation.

\begin{table}[t]
    \centering
    \setlength{\tabcolsep}{6pt}
    \caption{Average, maximum, and 95th-percentile solve time per step for BK-MPC and NMPC over 100 Monte Carlo scenarios.}
    \label{tab:solve_time}
    \begin{tabular}{lccc}
        \toprule
        \textbf{Controller} & \textbf{Average [s]} & \textbf{Max [s]} & \textbf{95\% [s]} \\
        \midrule
        BK-MPC & \textbf{0.0036} & \textbf{0.0088} & \textbf{0.0066} \\
        NMPC   & 1.154 & 1.600 & 1.3619 \\
        \bottomrule
    \end{tabular}
\end{table}

%%%%%%%%%%%%%%%%%%%%%%%%%%%%%%%%%%%%%%%%%%%%%%%%%%%%%%%%%%%%%%%%%%
%%%%%                       CONCLUSIONS               %%%%%%%%%%%%
%%%%%%%%%%%%%%%%%%%%%%%%%%%%%%%%%%%%%%%%%%%%%%%%%%%%%%%%%%%%%%%%%%

\section{CONCLUSIONS}\label{sec:conclusion}
This work presented a data-driven framework for collision-aware motion planning of mobile robots using Koopman operator theory. Starting from the unicycle kinematic model, nonlinear dynamics and obstacle avoidance constraints were lifted into a higher-dimensional observable space, where a bilinear EDMD approximation was identified. The bilinear realization was shown to accurately capture both system dynamics and quadratic terms from collision-avoidance constraints, outperforming linear Koopman realizations that failed to represent state–input couplings.

A convex QP-MPC formulation was then derived in the lifted space by freezing the bilinear terms at each prediction step, enabling computationally efficient trajectory optimization with moving obstacles. The proposed BK-MPC framework demonstrated accurate path planning and safe obstacle avoidance, while reducing the average computation time by 320 times compared to an NMPC formulated and solved in the original state space, thus ensuring feasibility for real-time applications on embedded hardware.

The results demonstrate the potential of bilinear Koopman models as a compelling alternative to conventional nonlinear model predictive control, enabling global linearization of nonlinear dynamics and constraints while retaining the tractability of linear MPC formulations. Future work will extend this framework to more complex vehicle dynamics, multi-agent interactions, and experimental validation on robotic platforms.

% \end{thebibliography}
\bibliographystyle{IEEEtran}
\bibliography{refs}

\end{document}